\documentclass[runningheads]{llncs}
\usepackage{graphicx}
\usepackage{subcaption}
\begin{document}
\title{SIR: Similar Image Retrieval for Product Search in E-Commerce}
%
%
\author{Theban Stanley\inst{1} \and
Nihar Vanjara\inst{1} \and
Yanxin Pan\inst{1} \and Ekaterina Pirogova\inst{1} \and Swagata Chakraborty\inst{1} \and Abon Chaudhuri\inst{1}}
\authorrunning{T. Stanley et al.}
%
\institute{Walmart Labs, Sunnyvale CA, USA 
\email{\{theban.stanley,yanxin.pan,achaudhuri\}@walmartlabs.com}}
\maketitle              
\begin{abstract}
We present a similar image retrieval (SIR) platform that is used to quickly discover visually similar products in a catalog of millions. Given the size, diversity, and dynamism of our catalog, product search poses many challenges. It can be addressed by building supervised models to tagging product images with labels representing themes and later retrieving them by labels. This approach suffices for common and perennial themes like ``white shirt" or ``lifestyle image of TV". It does not work for new themes such as ``e-cigarettes", hard-to-define ones such as ``image with a promotional badge", or the ones with short relevance span such as ``Halloween costumes". SIR is ideal for such cases because it allows us to search by an example, not a pre-defined theme. We describe the steps - embedding computation, encoding, and indexing - that power the approximate nearest neighbor search back-end. We also highlight two applications of SIR. The first one is related to the detection of products with various types of potentially objectionable themes. This application is run with a sense of urgency, hence the typical time frame to train and bootstrap a model is not permitted. Also, these themes are often short-lived based on current trends, hence spending resources to build a lasting model is not justified. The second application is a variant item detection system where SIR helps discover visual variants that are hard to find through text search. We analyze the performance of SIR in the context of these applications.
\end{abstract}
\section{Introduction}
\label{sec:intro}
Product data in catalogs owned by online retailers consist of text (title-description etc.), key-value pairs (attributes), and images. A number of internal systems and customer-facing applications leverage images to search and discover product(s) of interest. In some cases, the inherent nature of the application warrants a search through images. Also, the results of an image search usually complement that of a text search in most use cases. 

In this paper, we present a visual similarity-based product search and retrieval system built and deployed to address a number of business use cases at Walmart. Image search has come a long way with the recent advances in deep learning. However, building such a system that has the ability to scan through millions of images in a few seconds is a challenging task. The deep learning based fingerprints (or embeddings) created from images contain rich and complex information, but creating them is a compute-intensive task until GPUs are available in excess. Creating a search index on top of such large floating point arrays is not straightforward either. We present in this paper our process of encoding the embeddings so that they lend themselves well to popular search indexes like Elasticsearch and can be used to retrieve approximate nearest neighbors.

Our system is currently used in two business-critical applications. In this application-focused paper, we highlight how image similarity search plays a central role such applications. 
\begin{itemize}
    \item \textbf{Offensive or non-compliant product search}: The quality and compliance of our catalog are maintained through scheduled and on-demand searches for potentially offensive products. This discovery process demands a quick turnaround which makes the path of building supervised classifiers unattractive. Rather, a search tool that would accept one or a few known examples as a query and return more products with similar images is needed. 
    \item \textbf{Variant grouping}: In this classic e-commerce problem, items varying by color, size etc. are grouped together and presented to the customer at once on a single page. To create such groups from the catalog, we often start with a seed item and try to limit the search space to a pool of similar items. This pool of similar candidates can be created by text or image search or both. Our experiments suggest that image search often retrieves candidates that complement the ones retrieved by text search. 
\end{itemize}
Our system SIR has the potential to be used in other applications as well. With visual exploration emerging as an upcoming trend in retail, our image search index based on catalog product images can eventually become the back-end of a customer-facing visual search system. Also, SIR is used by data scientists to augment their training datasets with similar images. They often deal with machine learning problems where the data distribution across classes is highly skewed. This tool helps find training examples for poorly represented classes.

We optimize SIR for two objectives: search accuracy and query performance. We achieve high search accuracy by finding the most optimal deep learning based embedding after examining a few candidates. We achieve near real-time performance by encoding and indexing those embeddings in a scalable manner. The following sections of the paper delve into the technical details of the system and showcase its performance with appealing case studies of real applications.

\section{Related Work}
\label{sec:relwork}
In recent years, content-based image retrieval from large data sets has bifurcated into two distinct approaches.  Systems like FAISS~\cite{faiss} and NMSLib~\cite{malkov2016efficient} treat embeddings as first class citizens. At the time of this writing, these systems are typically scaled vertically by taking advantage of GPU based parallelisms. On the other hand, the older, mature search systems like Elasticsearch and Solr come with built-in support for scaling text-based searches to millions of documents. The above two approaches have been empirically compared by Mu et al.~\cite{comp}. As combining image and text searches(multi-modality) was an integral part of our overall solution, we decided on leveraging the second approach. Also, given the size of our catalog, a distributed system with in-built sharding was preferred. 

Many state-of-the-art image retrieval systems rely on very high dimensional features, known as \textit{embeddings} extracted either from a pre-trained network or by fine-tuning a deep neural network~\cite{deepcode}. Our system has experimented with a number of popular models such as VGG16~\cite{vgg16}, Resnet50-v2~\cite{resnet50}, Inception-v2~\cite{inceptionv2} and EfficientNet~\cite{effnet}. Deep learning based hashing, binarizing or a combination of them~\cite{dhn,dvsq,dch} are applied to the embeddings to reduce their storage cost and to improve the retrieval performance of indexes built on them.

From core functionality perspective, our system is a close neighbor to the visual search systems developed by various e-commerce companies~\cite{alibaba,ebay,pinterest}. However, a very important difference between those and SIR is that we apply our system to internal stakeholders; hence, the user interaction flow and other design choices are optimized for them. The actions taken with our system's results are very different from that of customer-facing visual search platforms.

\section{Technical Details} 
\label{sec:tech}
\begin{figure}[t]
\includegraphics[width=\textwidth]{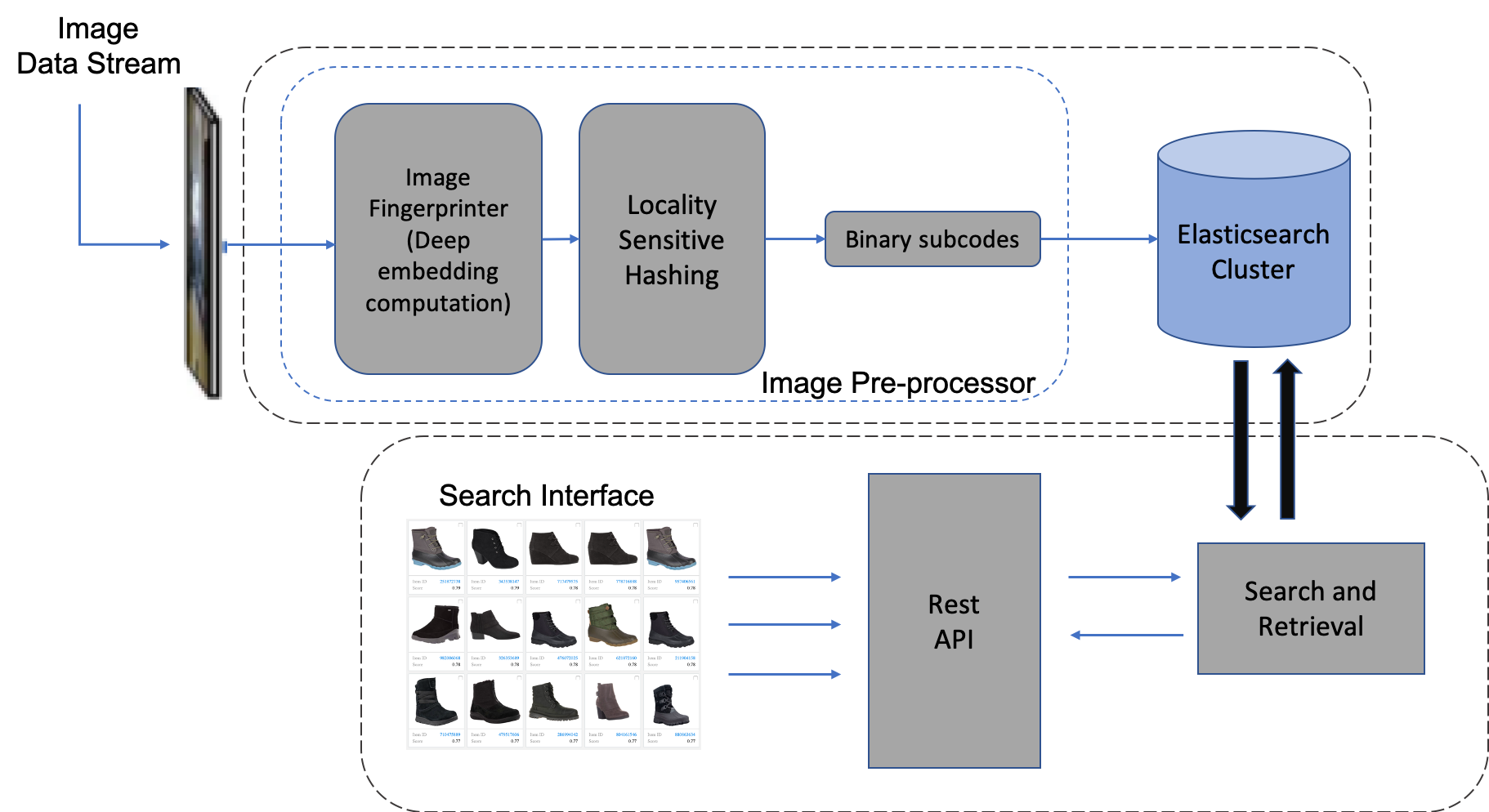}
\caption{System architecture of similar image retrieval tool (SIR). The top block outlines the back-end process of computing, encoding, and indexing embeddings. The bottom block shows the search interface.} \label{overall}
\end{figure}

The core of the system (Figure~\ref{overall}) revolves around fingerprinting every image to capture salient features and persisting it in a search index that would allow for efficient search and retrieval of nearest neighbors. We have the ability to use shallow fingerprinting techniques like phash~\cite{phash} or deep learning based embeddings~\cite{vgg16}. Most of our use cases require the ability to be invariant to slight changes in the image including positional and rotational variations. Also, the deeper and semantic aspects to embedding based fingerprinting was preferred.
\subsection{Embedding Generation}
In this step, we convert each image into a \textit{fingerprint} or \textit{signature} or \textit{unique descriptor}. Under the hood, the fingerprints are essentially embeddings computed from a suitable deep neural network. We have experimented with a large number of techniques for embedding generation and settled down on VGG16 as our primary network. The embeddings are taken off the final fully connected layer of VGG16.
\begin{figure}[t]
\includegraphics[width=\textwidth]{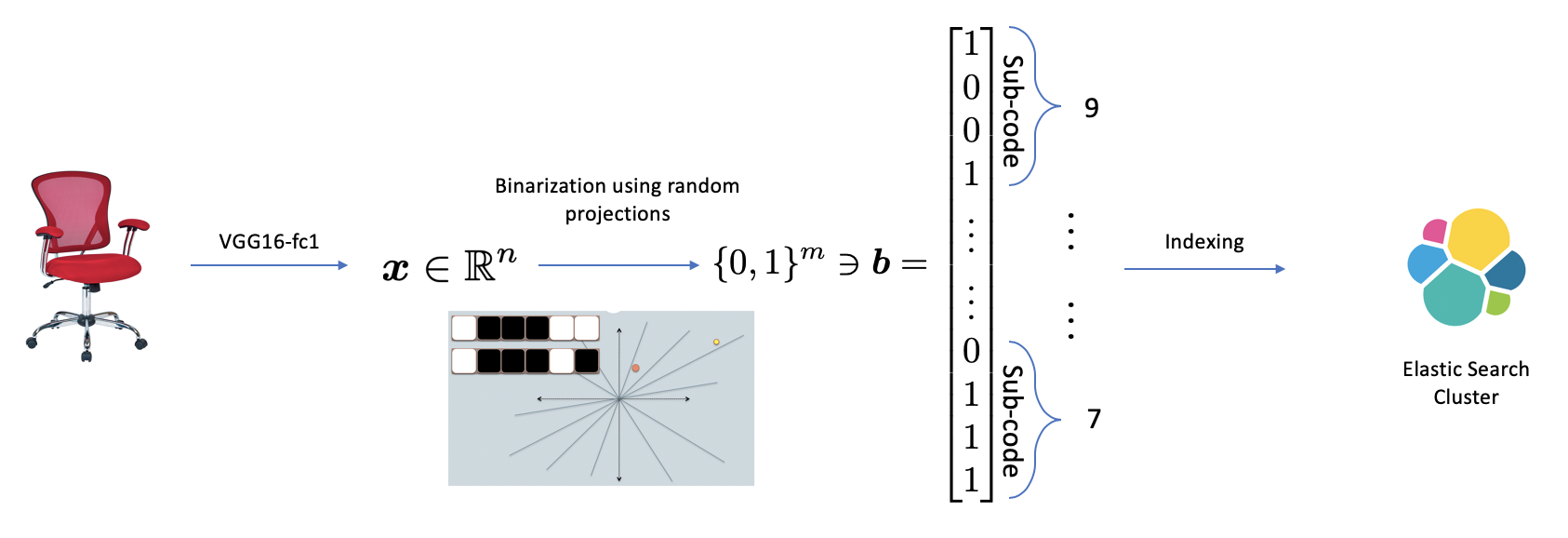}
\caption{Process of generation of subcodes from an image embedding followed by index creation on Elasticsearch.} 
\label{fig:index}
\end{figure}
\subsection{Index Creation}
A typical embedding is a high dimensional vector consisting of floating point numbers. At search time, both the high dimensionality and the need to numerous floating point comparisons are big hindrances to a scalable, near real-time implementation. In order to solve this, we employ a variant of locality sensitive hashing (LSH)~\cite{Gionis99similaritysearch} that binarizes the embedding vector by partitioning the embedding space using random projections. By constraining the dimensionality of this new binarized space, we can also mimic the effects of dimensionality reduction on the scalability of the search system. In its binarized form, the embeddings are further split into smaller subcodes~\cite{mu2019fast} and ingested into the Elasticsearch index (Figure~\ref{fig:index}). The subcoding enables us to take advantage of the pigeonhole principle and enforce early abandonment of search candidates which in turn helps us achieve sub-linear search times~\cite{mu2019fast}. At retrieval time, we also take advantage of Elasticsearch's ability to compute efficient hamming distance calculations in the form of bit operations.

Our product catalog is an ever-changing system. The business applications focus mostly on the images that were added to the catalog in last few weeks. Hence, we have designed the index creation as a rolling process so that the new and recently updated images are always indexed. The current deployed system listens to a Kafka~\cite{kafka} topic that streams new and updated images. On receiving an image, we compute its embedding, transform it into the suitable format and store in Elasticsearch~\cite{elastic}. A rolling index (last 3 months) of newly created images is maintained for subsequent search and retrieval.

The rolling nature of the application makes hash-based indexing a preferable choice over techniques that learn representations collectively from a static dataset such as principal component analysis (PCA). As the catalog changes, the optimal principal components change as well, requiring frequent re-computation of them.  

\subsection{Image Retrieval}
In the retrieval phase, a query image (also called \textit{seed image}) is provided to the system through the front end. In the back end, the query image is converted into an embedding and its nearest neighbors are retrieved from the indexed store. The retrieved images are presented in a grid in order of similarity with the query. Each result image is shown with a checkbox, allowing the user to select only the relevant ones from the grid. 

\section{Applications}
SIR is designed as a generic image-based similarity platform. The analysis of the core algorithm can be found in Mu et al.~\cite{mu2019fast}. In this paper, we focus on two implementations of SIR deployed to address two business application. Its performance is a function of a number of factors including the data on which the index is created. Hence, we present the system's performance in the context of specific applications. 
\subsection{Non-compliant Product Detection}
\label{subsec:t_and_s}
In a large company like Walmart, it is a common practice to identify offensive themes in products and mark them on a regular basis. Given the size and the diversity of our catalog, this daunting task is akin to finding a needle in a haystack. Product search deals with two distinct types of  themes. The first one is characterized by well-defined requirements, with a decent availability of training data. Also, these themes are usually relevant throughout the year. Hence, we address this type of themes by building supervised models~\cite{optimal,logo}. The second type of themes is characterized by ill-defined requirements. They are usually volatile and relevant for a short period of time (e.g. unauthorized sale of products at a specific time of the year). Given the sense of urgency which they come with, training and bootstrapping a new model is often too slow. Also, only one or two examples are usually available, hence finding enough data for training a model is nearly impossible. SIR is an ideal solution for addressing these ephemeral themes. We have built and deployed a platform with SIR at the core to address such issues. The platform consists of following two modes of operation:

\begin{itemize}
\item \textbf{Streaming:} The new products that get added to our catalog need to be constantly monitored for various issues. We provide a version of SIR that leverages the image similarity technique described in Section~\ref{sec:tech} to quickly identify such issues in new products. We accomplish this by listening to triggers that are generated as new products get ingested into our catalog. We fingerprint each new product and store them in an Elasticsearch index. The index is engineered to have a rolling window (currently set at past 3 months) of new products.
\item \textbf{Full Catalog Scan}: In addition of checking new products, business often needs to scan large parts of the catalog to find products similar to an example at hand. For this purpose, we provide a portal where an analyst can define rules based on image and text. An example rule would be an image of an e-cigarette and a filter that says ``product title contains e-cigarette". We use these rules to fetch parts of the catalog and then scan them in more detail. The fingerprinting technique discussed in Section~\ref{sec:tech} is used in two distinct ways in this portal. \textbf{Simulation:} Given a rule, we first scan a rolling index of sampled products to provide real-time feedback on the effectiveness of a rule as it get defined by the analyst. This is accomplished by maintaining an index of a good representation of the catalog. The simulation results help the analyst fine tune her rule and also the similarity thresholds that would lead to expected precision and recall. \textbf{Sweep:} Once the simulation is done, a full-fledged scan of the catalog is triggered, preferably on a GPU cluster, where we stream and compare every product to the set of finalized rules defined by the analyst. Empirically, the combination of image fingerprinting and text based filtering has proven to be very effective in identifying and flagging offensive products.
\end{itemize}

\subsubsection{Analysis of Search Quality}
We present the precision-recall characteristics of the image similarity technique in the context of our trust and safety application.

Given an application, search quality of the image similarity technique is dependent on two factors: the model used to generate embeddings and the level of binarization. We have repeated the following experiment for five different types of embeddings. For each embedding type, we populated the Elasticsearch with the embeddings from 1.5 million images of top-selling products. In order to test, we also ingested 10000 offensive images that were related to 3600 query images. For each query image, we compared the retrieved results against the ground truth to compute three metrics: Mean R-Precision, Mean Average Precision@K and Recall.
\begin{itemize}
\item R-Precision~\cite{rprecision} is useful when the number of relevant images varies from query to query. For example, if R images are relevant to a query image, R-precision ($r/R$) is computed based only on the top R images returned by the system. Mean R-Precision is the average of R-Precisions of all the queries. %
\item Precision@K~\cite{irbook} is computed based on the first K returned results. Average Precision@K is the average of AP for 1 to K. Average precision takes into account the position of the relevant documents, making it very useful for measuring the quality of search systems. Mean average precision@K is the average of AP@K over all the queries.
\item We also compute approximate recall@1000 with the assumption that all the relevant documents are either returned within top 1000 or not. 
\end{itemize}
The results of this experiment is shown in Table~\ref{tab1}. As the table indicates, we achieved best search quality with VGG16 embeddings. We then experimented with two binarized variations of VGG16 to understand the impact of binarization on search quality. The results, presented in Table~\ref{tab2}, indicate that both MAP and R-Precision is impacted by only about $2\%$ with the subcoded embeddings.
\begin{table}[thb]
\centering
\caption{Comparison of SIR search quality for different embedding types}
\label{tab1}
\begin{tabular}{|l|l|l|l|l|l|}
\hline
Embedding Type & MAP@1 & MAP@5 & MAP@10 & Mean R-Precision & Approx. Recall\\
\hline
VGG16 & 0.993 & 0.79 & 0.779 & 0.827 & 0.989\\
Inception-v2 & 0.993 & 0.688 & 0.663 & 0.711 & 0.801\\
ResNet-50v2 & 0.993 & 0.774 & 0.761 & 0.81 & 0.986\\
EfficientNet-b4 & 0.995 & 0.592 & 0.576 & 0.62 & 0.631\\
Custom & 0.992 & 0.772 & 0.76 & 0.806 & 0.993\\
\hline
\end{tabular}
\end{table}

\begin{table}[htb]
\centering
\caption{Comparison of SIR search quality for different levels of subcodings}
\label{tab2}
\begin{tabular}{|l|l|l|l|l|l|}
\hline
Embedding Type & MAP@1 & MAP@5 & MAP@10 & Mean R-Precision & Approx. Recall\\
\hline
VGG16 & 0.993 & 0.79 & 0.779 & 0.827 & 0.989\\
VGG16 with 512 subcodes & 0.993 & 0.784 & 0.774 & 0.824 & 0.984\\
VGG16 with 256 subcodes & 0.993 & 0.772 & 0.758 & 0.806 & 0.979\\
\hline
\end{tabular}
\end{table}
\subsubsection{Analysis of Query Response Time:} We also study the trade-off between search quality and query performance in the context of the same application. As Table~\ref{tab3} indicates, The mean query time of VGG16 is higher than Inception-v3 and Resnet50-v2. However, the mean query time reduces by 20\% as we switch to 512 subcodes of VGG16 from original VGG16 embeddings. It reduces by a massive 70\% as we move to 256 subcodes of VGG16. Table~\ref{tab2} has shown that this performance gain has been achieved with less than 2\% reduction in precision and recall.
\begin{table}[tb]
\centering
\caption{Comparison of SIR Elastic Search Query times for different embeddings.}
\label{tab3}
\begin{tabular}{|l|l|l|l|l|}
\hline
Embedding Type & min time (ms) & max time (ms) & mean time (ms) & total time (hrs)\\
\hline
ResNet-50v2 & 4764 & 5608 & 5201.644 & 5.538\\
Custom & 4448 & 5920 & 5458.125 & 5.500\\
Inception-v3 & 4227 & 5552 & 4622.075 & 4.525\\
VGG16 & 4892 & 6117 & 5600.235 & 5.560\\
VGG16 with 512 subcodes & 4112 & 4827 & 4481.130 & 4.449\\
VGG16 with 256 subcodes & 1568 & 1862 & 1631.270 & 1.112\\
\hline
\end{tabular}
\end{table}
%

\subsubsection{Performance Improvement with Text Filters:} 
\begin{figure}[htb]
\centering
\includegraphics[width=0.8\textwidth]{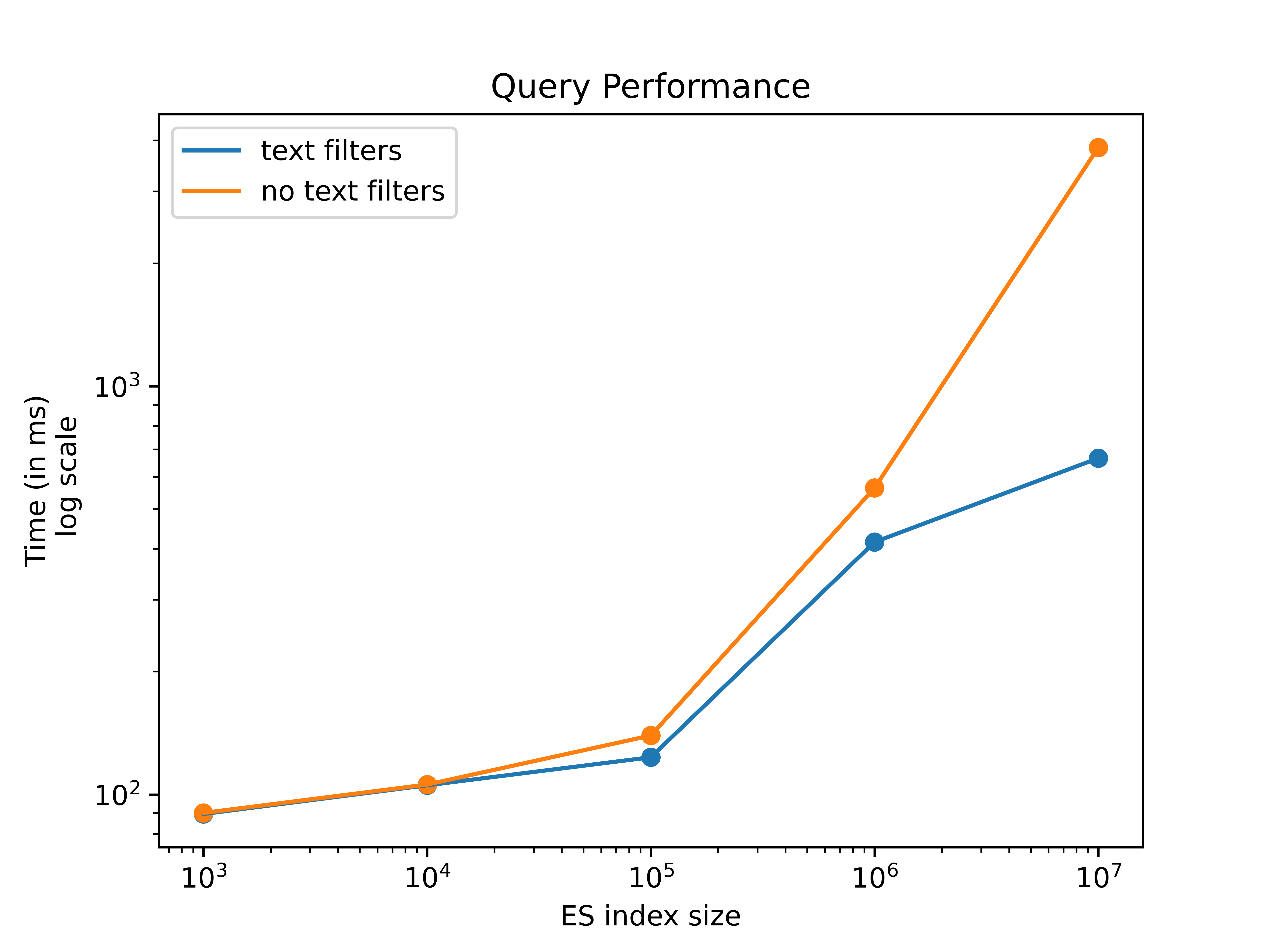}
\caption{Query performance on different index sizes with and without text based filtering} 
\label{fig:combined_index}
\end{figure}

In real applications, both image and text (mainly product title) contain information valuable for search. We have experimented with composite indexes. In this version, we create an image-based index as described earlier as well as a traditional Elasticsearch index of textual keywords. At the query retrieval phase, the image search space is first narrowed down by applying keyword-based text filters. Figure~\ref{fig:combined_index} shows the trends of query performance with and without the text filter. As the data size on which the indexes are built increases, the benefit of text filters on top of images becomes apparent. This early but promising result has opened up possibilities of turning this application into a multi-modal one.

\subsection{Variant Detection}
\label{subseec:var_det}
When shopping online, customers expect all item variants, for example, the same T-shirt in different sizes, on one item page so that they can easily make a well-informed shopping decision. Incorrect variant assortment in the item pages could result in poor customer experience and affect GMV due to an increased bounce rate as customers leave the site without any action. Traditionally, internal experts have been manually creating, consolidating, and updating the variant groups. This task is error-prone and time-consuming due to the volume of our catalog. To increase the variant grouping accuracy and efficiency, we developed a machine learning system to automatically generate variant suggestions so that experts only need to review a set of suggested variants instead of exploring the entire catalog. 

This variant grouping system consists of two stages: high-recall stage and high-precision stage. In the first stage, given a query product, a set of candidate variants is generated to narrow down the variant search space from the entire catalog to a few hundreds or thousands of products. In the second stage, high-precision classifiers are used to identify variants from the candidate set previously generated. In the first stage, a text similarity search was originally in place to retrieve candidates with similar product name and descriptions as the reference item. We deployed an implementation of the similar image retrieval (SIR) system to retrieve candidates that are visually similar to the reference item. Our hypothesis was that these two retrieval systems would fetch complementary variant candidates. 

\subsubsection{Performance Analysis:}
\label{sec:variant_perf}
\begin{figure}[htb]
\centering
\includegraphics[width=0.9\textwidth]{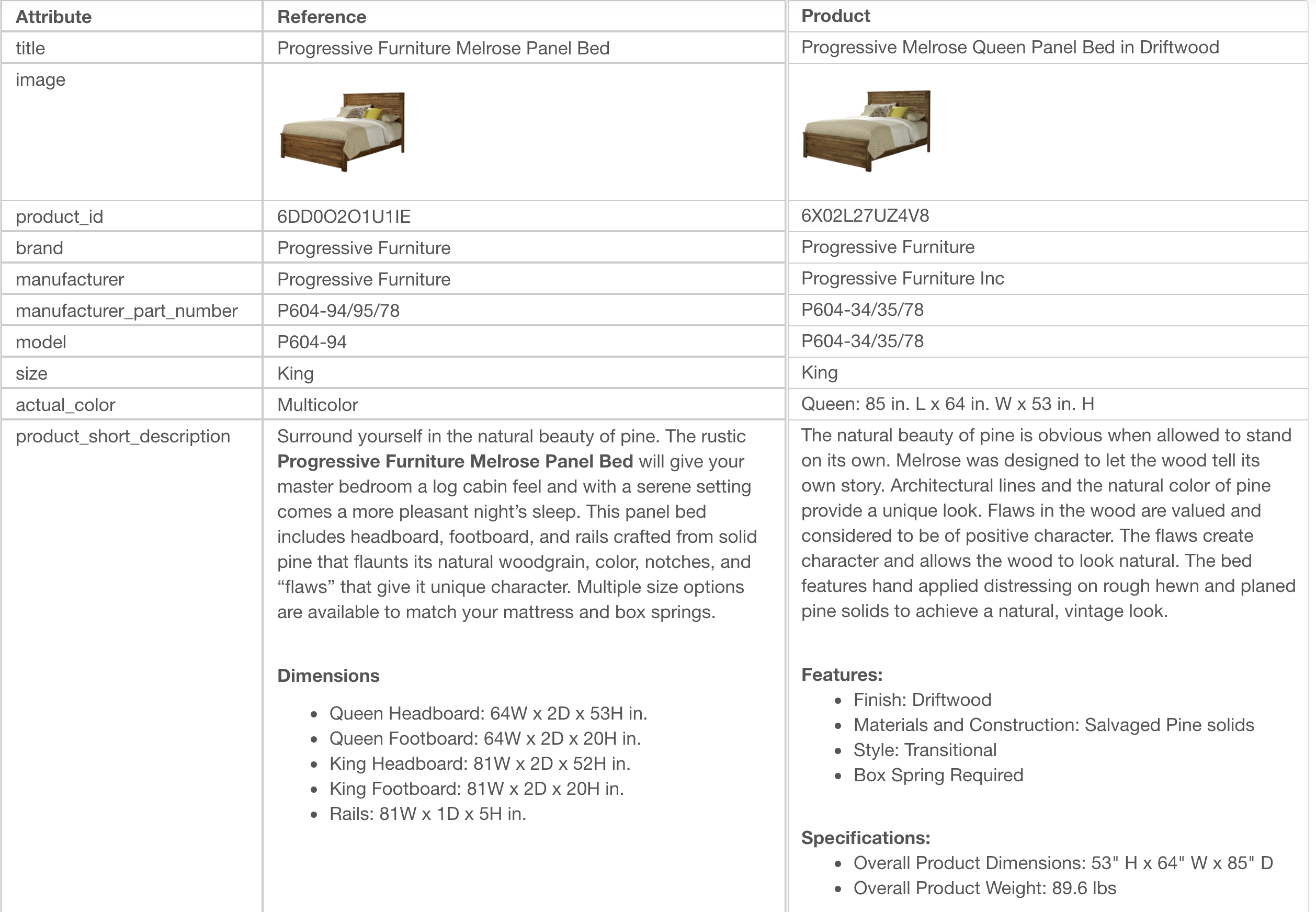}
\caption{The query product (on the right) and the variant candidate (on the left) have similar product image, but different product text information, especially the product descriptions. This variant candidate is only retrieved by SIR.} 
\label{fig_variant_example}
\end{figure}

To test our hypothesis, we measured the performance of the candidate generation system on a production-level dataset consisting of about 5,000 groups from thousands of product categories. For each reference item, we fetched about 1000 image and about 500 text based candidates (this discrepancy is due to the limitation of the library used to implement the text-retrieval system) independently and then combined them as well. It turned out that the recall based on the image-based candidates were already 13\% higher than that of the text-alone retrieval. The recall increased by 24\% after combining text and image-based candidate. Even with the discrepancy mentioned above, the numbers indicate that the image-based retrievals add significant value to the system. Figure \ref{fig_variant_example} shows an example where the variant is retrieved in the image SIR but is missed in text-alone retrieval. Though the text information for both products are semantically similar, the actual words, phrases, and writing style are so different that it is challenging for text-alone retrieval system. This challenge is prevalent in marketplace settings where multiple sellers for a single product are active. For such cases, product images are less subjective and harder to modify, hence image-base retrievals are critical.

\begin{figure}[htb]
\centering
\includegraphics[width=0.6\textwidth]{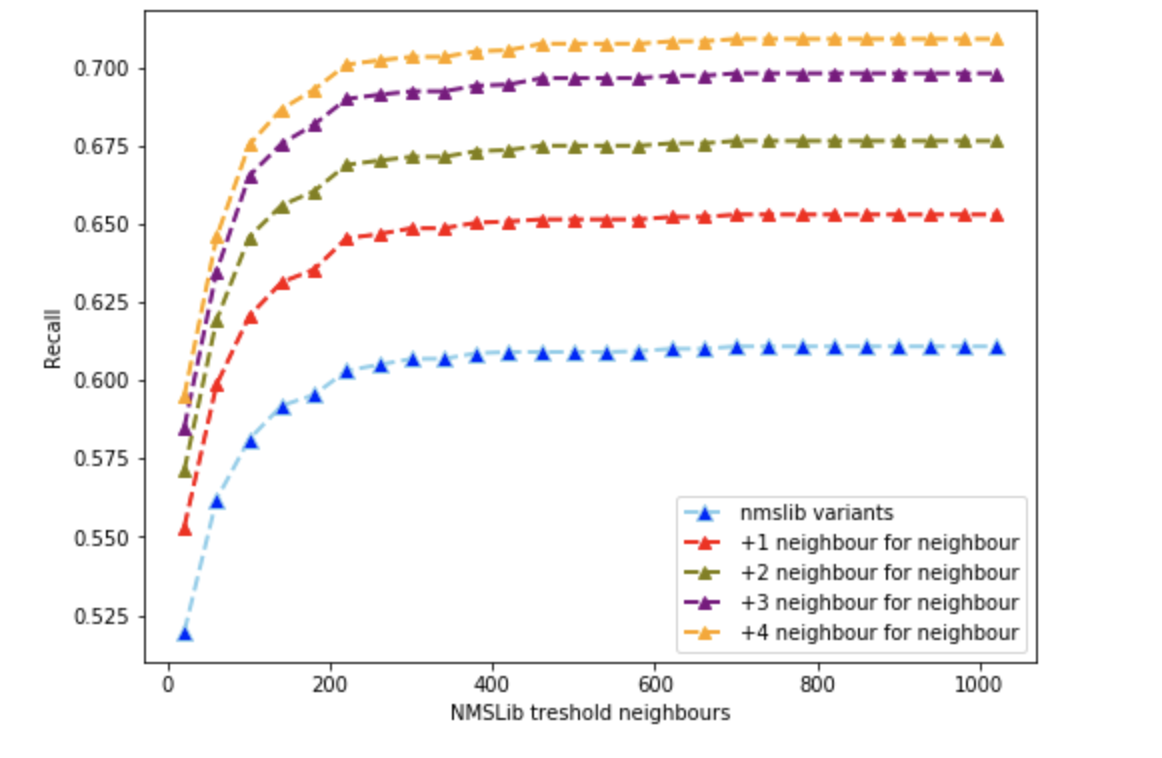}
\caption{Recall of any retrieval system saturates as we keep increasing number of candidates. Combining image-based retrieval with text elevates recall significantly over a text-alone system.} 
\label{fig_variant_sOfs_recall}
\end{figure}

\begin{figure}[htb]
\centering
\includegraphics[width=0.6\textwidth]{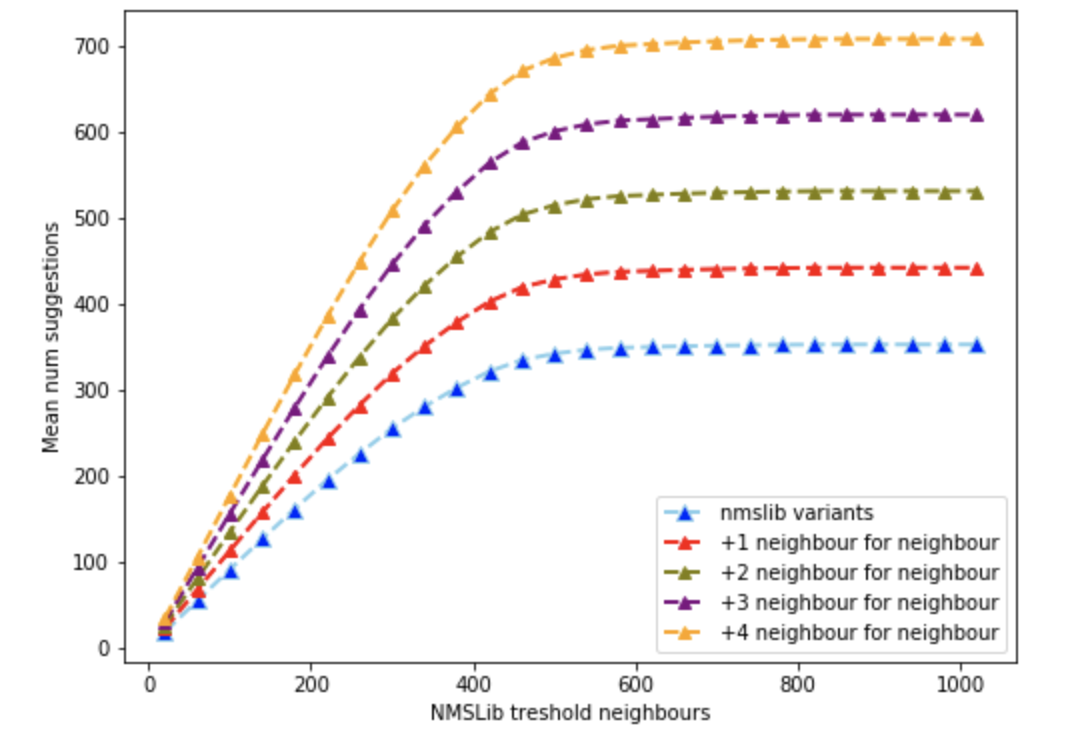}
\caption{Image and text together increases recall of text retrieval system with relatively small increase in number of candidates fetched.} 
\label{fig_variant_sOfs_num}
\end{figure}
We take a \textit{suggestion of suggestions} strategy where a few image candidates are retrieved for each text candidate. This strategy has two parameters: number of text-based candidates (shown as NMSLIB neighbors in the plots), namely $N$ and the number of image-based nearest neighbors for each text-based candidate, namely $K$. If we keep increasing $N$, a single retrieval system tends to saturate as the blue line shown in Figure \ref{fig_variant_sOfs_recall}. The recall increases little even though the number of searched items is doubled. Our experiment shows that a ``suggestion of suggestions" approach  can efficiently break through this saturation. Specifically, each retrieved text-based candidate  becomes the reference item for SIR. The top K neighbors retrieved in SIR are added into the candidate set. Figure \ref{fig_variant_sOfs_recall} shows how recall is increased for {k=1,2,3,4}. Figure \ref{fig_variant_sOfs_num} shows that this ``suggestion of suggestions" strategy significantly increase the recall that surpass the saturated point with insignificant increment in the number of fetched candidates. The y-axis of this plot shows the number of candidates averaged over all the queries used for testing.

\subsection{Visual Examples from Applications}
\begin{figure}[p]
\centering
\begin{subfigure}{\linewidth}
  \centering
  \includegraphics[width=.9\textwidth]{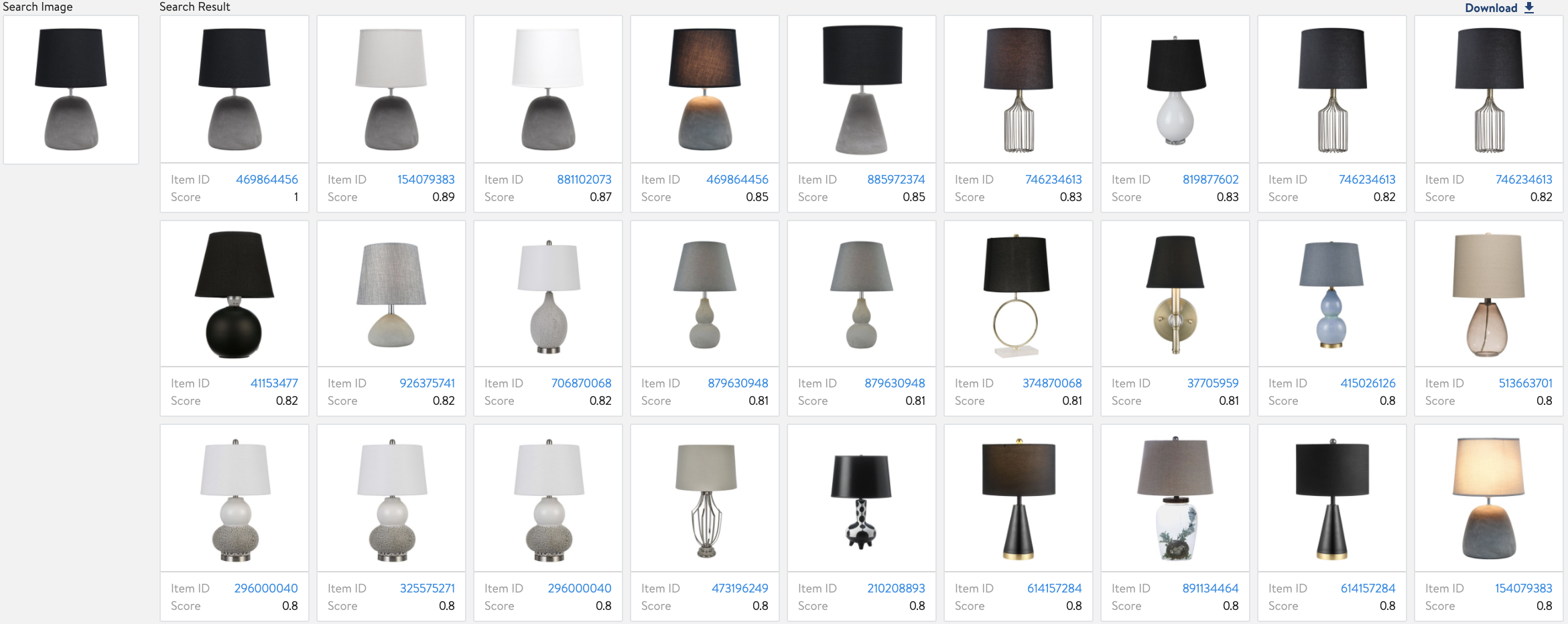}
  \caption{Discovery and retrieval of table lamps similar to the query.}
  \label{fig:sub1}
\end{subfigure}
\begin{subfigure}{\linewidth}
  \centering
  \includegraphics[width=.9\textwidth]{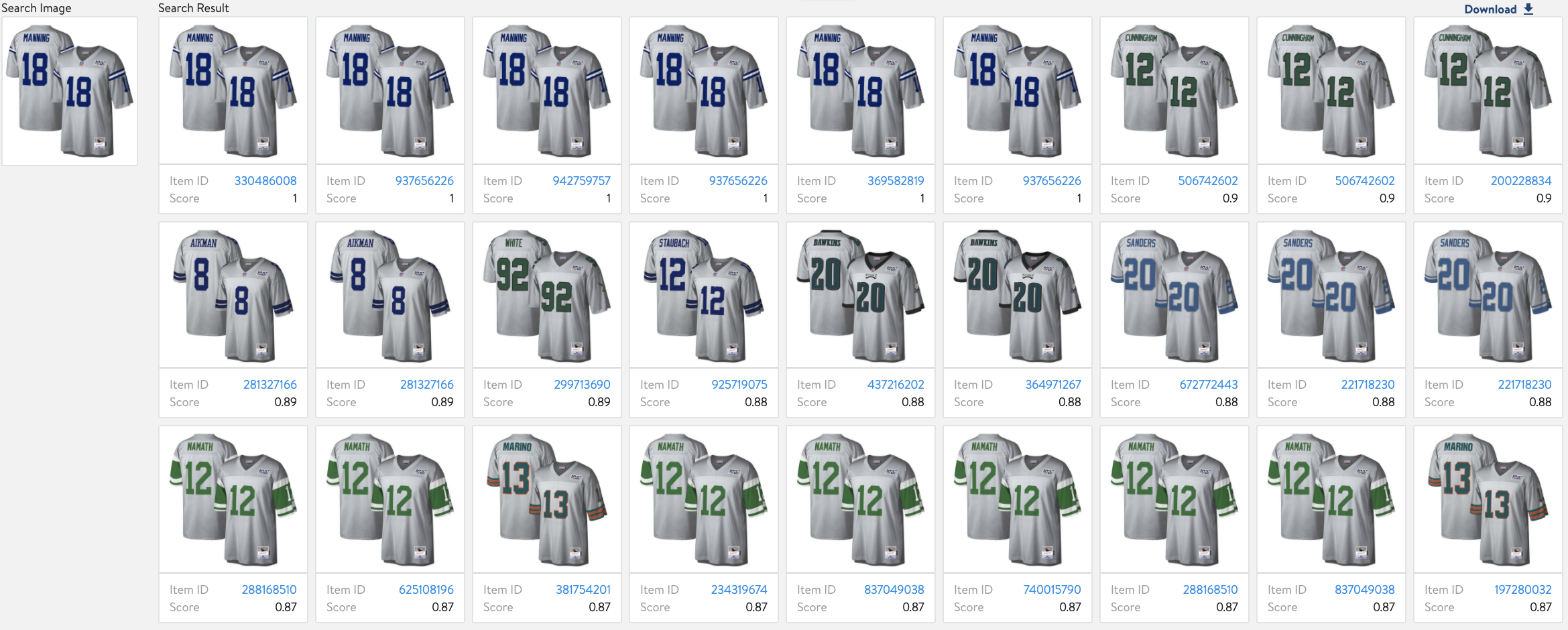}  
  \caption{Discovery and retrieval of variants of a sports t-shirt.}
  \label{fig:sub2}
\end{subfigure}
\begin{subfigure}{\linewidth}
  \centering
  \includegraphics[width=.9\textwidth]{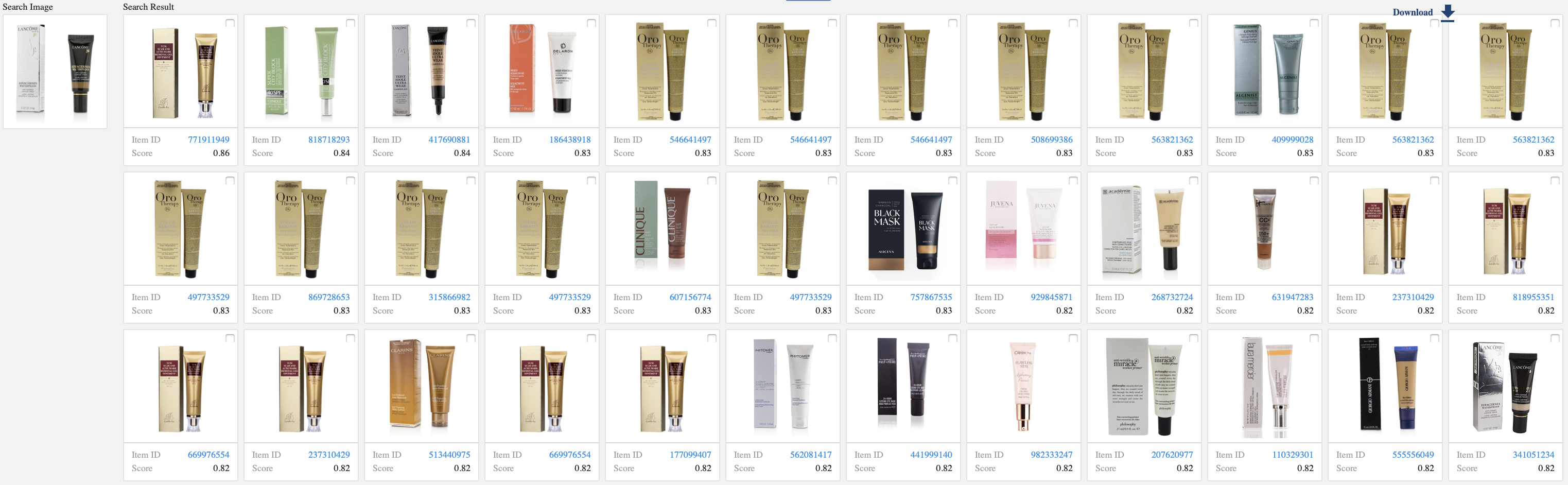} 
  \caption{Retrieval of cosmetic products that are likely to be variants from different brands.}
  \label{fig:sub3}
\end{subfigure}
\begin{subfigure}{\linewidth}
  \centering
  \includegraphics[width=.9\textwidth]{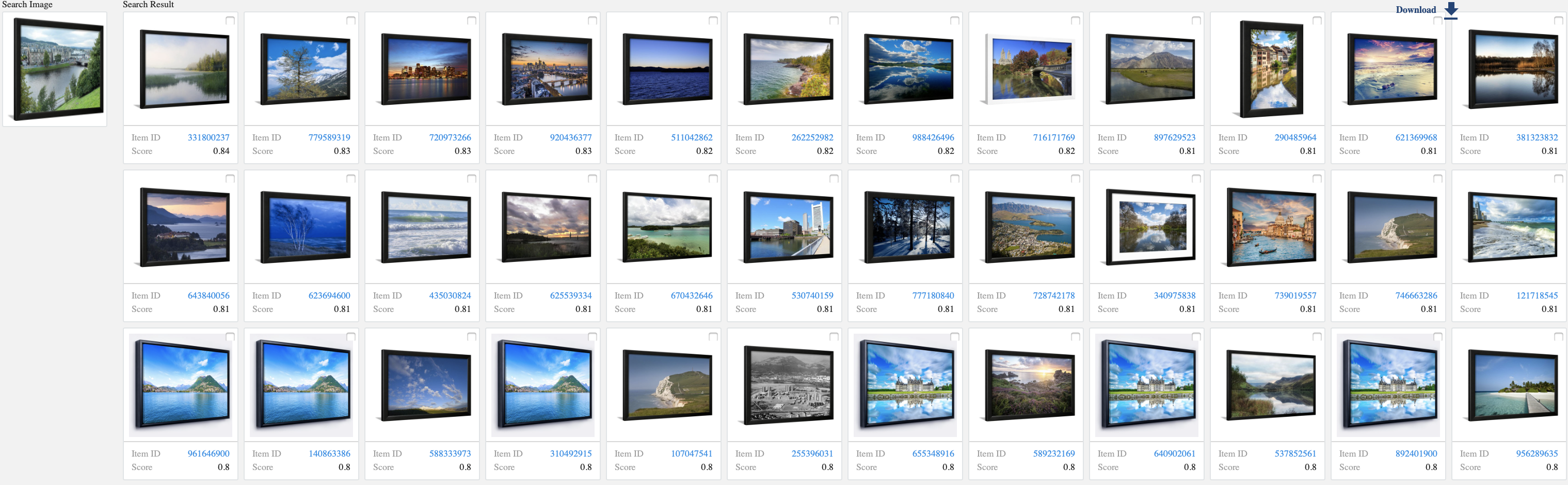}  
  \caption{Retrieval of tablets similar to an example}
  \label{fig:sub4}
\end{subfigure}
\caption{A number of examples demonstrating product search capability of SIR} 
\label{fig:sir_example1}
\end{figure}

Both the above mentioned applications regularly use SIR to discover products similar to a query example that is of interest to one of the users. The retrieval is based on an index of products that entered the catalog in the last one month or so. Figure~\ref{fig:sir_example1} showcases a few such examples and corresponding search results. 

The top one with a table lamp, Figure~\ref{fig:sub1}, shows how SIR retrieves similar products with subtle variations in shape, size, and color. Such variations either make them variants of the queried product or help in discovering products with a certain shape or style catering to the customer's choice.
Figure~\ref{fig:sub2} showcases a search for sports t-shirts where variations in the text on the t-shirt are captured. This is a relatively difficult example since our underlying embeddings were not trained to detect letters and numbers. However, SIR still identifies the graphic as a key feature and is able to fetch t-shirts with a similar graphic.

The third example in Figure~\ref{fig:sub3} is an attempt to find cosmetic products, most likely coming from different brands. All the results returned are near duplicates to the query, but not identical.

The last example in Figure~\ref{fig:sub4} retrieves tablet computers based on an example. This example actually highlights a limitation of our system. Since the underlying embeddings do not recognize the content being shown inside the screen of the tablet, it retrieved images primarily based on the overall shape of the object. If the user were looking for other tablets showing similar content on the screen, this search result would not satisfy her.
\section{Discussion and Future Work}
\label{sec:discussion}
The focus of this paper is on building systems using an embedding based image similarity algorithm~\cite{mu2019fast}. Hence, we present the performance of our system in the context of specific applications. Internal product datasets are used for the experiments. Even if the exact numbers change a bit when a similar system is built for another application, we are confident that the key insights will hold (such as the impact of sub-coding on precision-recall and query performance, or the benefit of text-based pre-filtering). We also skip the comparison between deep learning embeddings and conventional image hashes because there is enough evidence in the literature that conventional image hashes cannot perform nearly well beyond exact or near duplicated. 

The image search platform we have built and deployed is constantly undergoing improvements. On the algorithm side, we are experimenting on making the embeddings aware of regions of interest so that the users can submit queries with annotated regions of attention. We intend to upgrade the binarization of the embeddings to a learnable process using one of the deep hashing networks. can We also intend to scale the embedding computation and the image search using serverless compute offerings from various cloud enterprises. More involved text search is also underway.
\section{Conclusion}
\label{sec:conclusion}
In this paper, we present a similar image retrieval (SIR) tool designed and deployed to support a number of internal applications that need to discover products from Walmart's enormous product catalog. The system is developed by skillfully combining knowledge of deep learning, data management, and user experience. The core idea behind SIR can be used to build similar visual search tools for many other domains.  
%
%
%
\bibliographystyle{splncs04}
\bibliography{samplepaper}
\end{document}